\documentclass[9pt, conference]{IEEEtran}

\PassOptionsToPackage{dvipsnames,prologue}{xcolor}

\usepackage{caption}
\usepackage{subcaption}
\usepackage{type1cm}
\usepackage{lettrine}
\usepackage{amsmath}
\usepackage{listings}
\usepackage{footnote}
\usepackage{graphics}
\usepackage{multirow}
\usepackage{svg}
\usepackage{tabularx}
\usepackage{doi}
\usepackage{booktabs}

\usepackage{hyperref}

\usepackage{ifthen}
\usepackage[normalem]{ulem} 
\usepackage{xcolor}
\usepackage{amssymb}  

\newboolean{showedits}
\setboolean{showedits}{true} 
\ifthenelse{\boolean{showedits}}
{

	\newcommand{\del}[1]{\textcolor{red}{\sout{#1}}}
	
}{

	\newcommand{\del}[1]{}
	
}

\newboolean{showcolors}
\setboolean{showcolors}{true} 
\ifthenelse{\boolean{showcolors}}
{

}{

}

\newboolean{showcomments}
\setboolean{showcomments}{true}

\ifthenelse{\boolean{showcomments}}
{\newcommand{\nbc}[3]{
 {\colorbox{#3}{\bfseries\sffamily\scriptsize
 	\textcolor{white}{#1}}}
 {\textcolor{#3}{\sf\small$\blacktriangleright
 	${#2}$\blacktriangleleft$}}}
 }
{\newcommand{\nbc}[3]{}}

\definecolor{ibcolor}{rgb}{0.4,0.6,0.2}

\definecolor{metacolor}{rgb}{0.5,0,0.5}

\definecolor{ideacolor}{rgb}{1.0,0,0.5}

\definecolor{todocolor}{rgb}{0.9,0.1,0.1}

\definecolor{qcolor}{rgb}{0.2,0.0,0.9}

\definecolor{owcolor}{rgb}{0.2,0.4,0.2}

\definecolor{rkcolor}{rgb}{0.2,0.0,0.9}

\definecolor{rkcolor}{rgb}{0.2,0.0,0.9}

\begin{document}

\title{Architectural Implications of Neural Network Inference for High Data-Rate, Low-Latency Scientific Applications}

\author{
\IEEEauthorblockN{Olivia Weng \\ Alexander Redding}
\IEEEauthorblockA{UC San Diego}
\and
\IEEEauthorblockN{Nhan Tran}
\IEEEauthorblockA{Fermi National Accelerator Laboratory}
\and
\IEEEauthorblockN{Javier Mauricio Duarte\\ Ryan Kastner}
\IEEEauthorblockA{UC San Diego}
}

\maketitle
\begin{abstract}
With more scientific fields relying on neural networks (NNs) to process data incoming at extreme throughputs and latencies, it is crucial to develop NNs with all their parameters stored on-chip.
In many of these applications, there is not enough time to go off-chip and retrieve weights.
Even more so, off-chip memory such as DRAM does not have the bandwidth required to process these NNs as fast as the data is being produced (e.g., every 25 ns). 
As such, these extreme latency and bandwidth requirements have architectural implications for the hardware intended to run these NNs: 1) all NN parameters must fit on-chip, and 2) codesigning custom/reconfigurable logic is often required to meet these latency and bandwidth constraints. 
In our work, we show that many scientific NN applications must run fully on chip, in the extreme case requiring a custom chip to meet such stringent constraints. 
\end{abstract}
\section{Introduction}
As neural networks (NNs) become increasingly capable, disciplines outside of computer science are looking to NNs to play a part in their research. 
This is particularly prevalent in many scientific experiments, including particle physics, electron microscopy, and X-ray diffraction~\cite{duarte2022fastml}. 
As the scientific instruments used in these experiments become more advanced, they are able to produce data at extremely high rates, e.g., 40 TB/s~\cite{duarte2022fastml}.
Many of these fields are grappling with how to best collect and process the data to conduct the latest research at such extreme rates.
Several are using NNs to filter out data for post-processing later, as traditional algorithms have reached their limits~\cite{di2021reconfigurable}.
NNs are typically built in a brute force way, where researchers add more and more parameters to increase performance, scaling them out to enormous sizes, e.g., GPT-3 has 175 billion parameters, which amounts to 350 GB of data when stored with float16 precision~\cite{brown2020language}. 
However, this is not practical when data is coming in at rates of several TB/s because current computer architectures cannot execute such enormous NNs in time without being extremely expensive and impractical. 
As a result, researchers in these scientific fields are building edge NNs that can be executed on a machine within the computation budget ($O(\text{ns})$) allowed by the data rate.
This has several implications for the computer architectures that need to meet these extreme low-latency requirements:
\begin{enumerate}
    \item All NN parameters must fit on-chip.
    \item Fully on-chip inference often requires hardware-software codesign with custom/reconfigurable logic to meet latency and bandwidth constraints.
\end{enumerate}

In this paper, we argue for these architectural implications, evaluating on the LHC sensor benchmark, which exhibits a 40 TB/s data rate, as a case study.
From our results, we can see that on a benchmark with the most extreme data rate and latency requirements, on-chip inference and custom architectural NN implementation is the necessary path forward.

\begin{figure}[t]
\centering
    \includegraphics[width=\columnwidth]{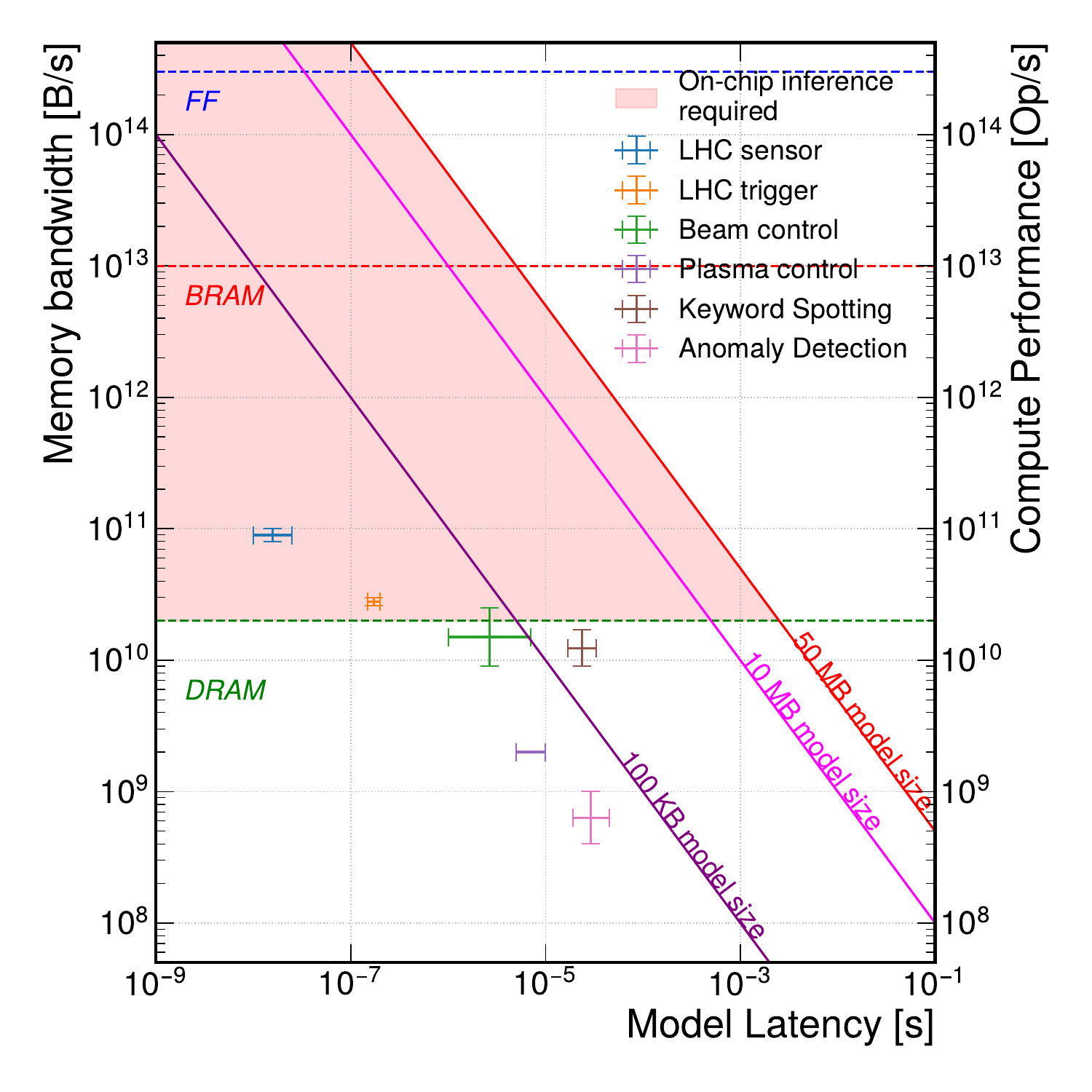}
    \caption{Many scientific and edge NNs must process incoming data at a high rate, requiring on-chip inference to process the data at least as fast as it arrives~\cite{duarte2022fastml, wei2023low, borras2022open}. This leads to extreme low-latency and high-bandwidth requirements.}
\label{fig:sciml}
\end{figure}
\section{Architectural Implications}

\subsection{All NN parameters must fit on chip}
Given the extreme bandwidth and latency requirements of many scientific machine learning tasks~\cite{duarte2022fastml, wei2023low}, on-chip memories such as flip-flops (FFs) and block random access memories (BRAMs) are often the only kinds of memory with both the bandwidth and latency capable of meeting these constraints. 
To demonstrate, we present~\autoref{fig:sciml}. 
We chart several scientific and IoT NNs according to their memory bandwidth for reading NN weights and the latency they achieve on field-programmable gate arrays (FPGAs) and application-specific integrated circuits (ASICs), focusing first on latency on the x-axis and memory bandwidth on the left y-axis.
For example, the LHC sensor NN requires 80 GB/s memory bandwidth since it loads 2000 1-byte weights every 25 ns.
This NN has bandwidth requirements that exceed the capabilities of DRAM (assuming DDR4's 20 GB/s bandwidth noted by the green horizontal dashed line).
It is therefore infeasible to access weights off-chip from DRAM.
As a result, the LHC sensor NN must rely on on-chip memory technology such as BRAM (red dashed line) and FFs (blue dashed line), which have significantly higher bandwidth at the expense of smaller memory capacity (\autoref{tab:mem}).

Next, let us focus on latency on the x-axis and compute performance (in operations/s) on the right y-axis to understand the diagonally charted lines.
To start, we make some assumptions.
First, we assume the NNs for each task are fully connected and the weights are 1 byte wide, as scientific benchmarks often work with heavily quantized, fully connected layers.
Also, these NNs process data with batch size 1, as data is usually streamed in one sample at a time.
Second, we assume these NNs execute on a performant FPGA, e.g., a Xilinx VU13P, using 5\,000 digital signal processors (DSPs) operating at 200 MHz, performs 1 TOP/s.
Third, we assume that with 1-byte weights and 1 TOP/s, we can handle 1 TB/s of bandwidth at a compute performance of 1 TOP/s. 
Based on these assumptions, consider a model with 10 MB of weights (the diagonal pink line). 
Executing this model at a latency of $10^{-5}$s requires 1 TOP/s with a memory bandwidth of 1 TB/s (left y-axis). 
From these assumptions, we can project how much compute performance and memory bandwidth a 10 MB model needs to run at different latencies. 
We also project the compute and bandwidth for a 100 KB model (purple diagonal line) and a 50 MB model (red diagonal line). 
Assuming that on-chip storage can only store a maximum of 50 MB of weights, we shade the region of this chart where on-chip inference is required in red. 
The floor is the maximum DRAM bandwidth and the ceiling is the 50 MB model's memory bandwidth and compute performance.

As such, NN implementations that fall in this shaded region must be executed with weights fully on-chip, as demonstrated by the LHC sensor, LHC trigger, and beam control NNs.
NN implementations that reside outside this region can instead rely on off-chip memory, e.g., the Anomaly Detection IoT NN.

Many of these scientific NN implementations will likely become more bandwidth-intensive as scientific instruments continue to advance. 
This implies that the NNs charted in \autoref{fig:sciml} will continue to move up and to the left.
Thus, on-chip NN inference will continue to persist in the future.

\begin{table}[t]
    \centering
    \begin{tabularx}{\columnwidth}{Xrrr}
        \toprule
         & \textbf{DRAM} & \textbf{BRAM} & \textbf{FF}\\
        \midrule
        Latency & 10 - 100 ns & 1 - 1.4 ns & 1 - 1.4 ns \\
        Bandwidth & 10s GB / s & TB / s & 100s TB / s \\
        Total capacity & GBs & MBs & 100s KBs \\
        \bottomrule
    \end{tabularx}
    \caption{On-chip memory storage, i.e., BRAMs and FFs, offers significantly more bandwidth and speed, making it the ideal option for our high-bandwidth, low-latency benchmarks~\cite{kastner2018parallel}. The tradeoff, however, is total capacity, as these on-chip options can only store megabytes of data as opposed to DRAM's gigabytes.}
    \label{tab:mem}
\end{table}

\subsection{Fully on-chip inference often requires hardware-software codesign}
To perform on-chip NN inference fast enough, researchers have developed a couple different architectures, such as spatial dataflow style and lookup table (LUT) based NN implementations, which heavily rely on hardware-software codesign to meet accuracy and hardware performance requirements.
Spatial dataflow style architectures, e.g., hls4ml~\cite{Duarte:2018ite} and FINN~\cite{umuroglu2017finn}, often target FPGAs but also target ASICs~\cite{di2021reconfigurable}.
These architectures implement the NN layer-wise, dedicating fabric to each layer, dataflowing them to achieve high throughput.
To get NNs to fit entirely on chip, these designs rely on quantizing the weights and altering the NN topology to achieve good accuracy and hardware performance.
An even more extreme architecture implements NNs entirely using LUTs~\cite{umuroglu2020logicnets}. 
Rather than implementing multiply-accumulates (MACs), the NN is executed entirely by using lookups.
However, since LUT-based NNs scale up exponentially ($O(2^n)$) with respect to neuron fan-in, extreme sparsity and quantization is required, so this method also leans heavily on hardware-software codesign techniques to construct, train, and implement a performant NN.

\textit{\underline{LHC Sensor Case Study}}: 
We compare the performance of the LHC sensor NN~\cite{di2021reconfigurable} implemented across several different hardware platforms to demonstrate how for a task with a extreme requirements (80 GB/s memory bandwidth and 25 ns latency budget) custom hardware is needed.
As seen in \autoref{tab:study}, only the ASIC can meet the 25 ns latency budget.
Our FPGA design was implemented using hls4ml~\cite{Duarte:2018ite}. The results demonstrate that for one of the most latency-sensitive scientific applications (LHC sensor), we can achieve similar performance to an ASIC by exercising hardware-software codesign on an FPGA. This relationship is not as feasible when developing for the traditional high-performance GPU architectures we evaluted against.

\begin{table}[t]
    \centering
    \resizebox{\columnwidth}{!}{%
    \begin{tabular}{lrr}
        \toprule
        \textbf{Hardware Platform} & \textbf{Latency (ns)} & \textbf{Throughput (inf/s)} \\
        \midrule
        ASIC~\cite{di2021reconfigurable} & 25 & 40\,000\,000  \\
        AMD ZCU104 FPGA & 37.5* & 26\,666\,667  \\
        Nvidia Jetson GPU (batch size = 1) & 17\,456 & 52\,185  \\ 
        Nvidia Jetson GPU (batch size = 512) & 47\,851 & 10\,343\,270  \\ 
        Nvidia V100 GPU (batch size = 1) & 10\,253 & 43\,185  \\
        Nvidia V100 GPU (batch size = 512) & 17\,456 & 22\,246\,912  \\
        CPU & 1\,800\,000 & 550  \\
        \bottomrule
    \end{tabular}%
    }
    \caption{\textbf{LHC sensor NN benchmarked across hardware.} The ASIC outperforms all other platforms and is a must for meeting the 25 ns latency requirement~\cite{di2021reconfigurable}. * - This latency is from simulation and presents a best case.}
    \label{tab:study}
\end{table}

\section{Conclusion}
Our key takeaways are: 1) To meet the extreme low-latency and high bandwidth requirements of many scientific tasks, all NN parameters must fit on-chip. 2) To fit all NN parameters on-chip, hardware-software codesign on custom and reconfigurable hardware facilitates meeting the requirements of these extreme tasks.



%

\bibliographystyle{IEEEtran}
\bibliography{refs}

\end{document}